%% file: equall.tex
\documentclass{article} 
\usepackage{graphicx}        
\usepackage{booktabs}        
\usepackage{multirow}        
\usepackage{threeparttable}  
\usepackage{siunitx}         
\usepackage[table]{xcolor}   
\usepackage{makecell}        
\usepackage{array}           
\usepackage{pdflscape}       
\usepackage{adjustbox}       
\usepackage[most]{tcolorbox} 
\usepackage{microtype}       
\usepackage{hyperref}        
\usepackage{url}             
\usepackage{amssymb}         
\usepackage{float}           
\usepackage{CJKutf8}         
\usepackage{equall}   
\usepackage{longtable}  

\newcolumntype{P}[1]{>{\raggedright\arraybackslash}p{#1}}

\definecolor{CustomBlue}{RGB}{57,83,191}

\newcommand{\captable}{\mathcal{C}}
\newcommand{\captableref}{C^{\mathrm{ref}}}
\newcommand{\captablevirtual}{C^{\mathrm{virt}}}

\newtcbox{\clustertab}[1]{on line, box align=base, colback={#1},colframe={#1},size=fbox,arc=2pt,top=-1.5pt, bottom=-1.5pt, left=-1.5pt, right=-1.5pt, boxrule=0pt, enlarge left by=1pt}

\usepackage{arydshln}
\makeatletter
\def\adl@drawiv#1#2#3{%
        \hskip.5\tabcolsep
        \xleaders#3{#2.5\@tempdimb #1{1}#2.5\@tempdimb}%
                #2\z@ plus1fil minus1fil\relax
        \hskip.5\tabcolsep}
\newcommand{\cdashlinelr}[1]{%
  \noalign{\vskip 1.3pt
           \global\let\@dashdrawstore\adl@draw
           \global\let\adl@draw\adl@drawiv}
  \cdashline{#1}[.4pt/2pt]
  \noalign{\global\let\adl@draw\@dashdrawstore
           \vskip 3pt}}
\makeatother

\usepackage{enumitem}
\setlist[itemize,enumerate]{leftmargin=*}

\usepackage{graphicx}
\usepackage{pgf-pie}

\usepackage{pgfplots}
\usetikzlibrary{pgfplots.groupplots}
\pgfplotsset{compat=1.3}
\usepackage{tikz}
\usetikzlibrary{patterns}
\usetikzlibrary{shapes.geometric} 
\usepackage{caption}
\usepackage{subcaption}
\usepackage{wrapfig}  

\usepackage{colortbl}
\usepackage{xcolor}

\usepackage[most]{tcolorbox}

\author{
 Pierre Colombo$^*$, Malik Boudiaf$^*$, Allyn Sweet, Michael Desa, Hongxi Wang, Kevin Candra, Syméon del Marmol
  \vspace{-.01cm}
 \begin{center}
     \texttt{\{firstname\}@equall.com}
 \end{center}
}

\title{Does It Tie Out? Towards Autonomous Legal Agents in Venture Capital}
\abstract{

Before closing venture capital financing rounds, lawyers conduct diligence that includes \emph{tying out} the capitalization table: verifying that every security (e.g., shares, options, warrants) and issuance term (e.g., vesting schedules, acceleration triggers, transfer restrictions) is supported by large sets of underlying legal documentation. While LLMs continue to improve on legal benchmarks, specialized legal workflows, such as capitalization tie-out, remain out of reach even for strong agentic systems: the task requires multi-document reasoning, strict evidence traceability, and deterministic outputs that current approaches fail to reliably deliver. We characterize capitalization tie-out as an instance of real-world benchmark for legal AI, analyze and compare the performance of existing agentic systems, and propose a world model architecture towards tie-out automation—and more broadly as a foundation for applied legal intelligence. 
\footnotetext[1]{Equal contributions}

}
\begin{document}

\maketitle
\section{Introduction}
Verifying \textbf{legal ownership}  of a company is a bottleneck in private market transactions, from venture financings to M\&A deals. Before a deal closes, lawyers must manually reconcile thousands of pages against ownership records, a process commonly referred to as \emph{cap table tie-out} or capitalization due diligence. This work is critical to confirm financial allocations up to and arising from the transaction, while also tedious, error-prone, and typically performed under capped fees that compress margins. Large language models have shown impressive capabilities on legal reasoning benchmarks  \cite{comanici2025gemini,colombo2024saullm,colombo2024saullm1,katz2023gpt,jaech2024openai} and contract analysis tasks \cite{re2019developing,bhambhoria2024evaluating}, yet this recurring verification workflow remains categorically manual. This paper investigates why and what is required to bridge the gap.

In any financing or acquisition the disbursing party must verify all ownership claims, contractual obligations, potential liabilities, and regulatory risks before funds change hands \cite{boone2007firms}. This \textbf{legal risk assessment} centers on the \emph{dataroom}, a repository containing the company’s legal and financial history: incorporation filings, stock purchase agreements, option grants, convertible instruments, board consents, and their amendments. Market data shows over 25,000 venture capital financings per year, with circa 15,000 originating in the US alone. Routine Seed and Series A rounds generate datarooms with thousands of pages, and Series B and after often reach tens of thousands of pages.

\begin{figure}[ht]
    \centering
    \includegraphics[width=0.7\textwidth]{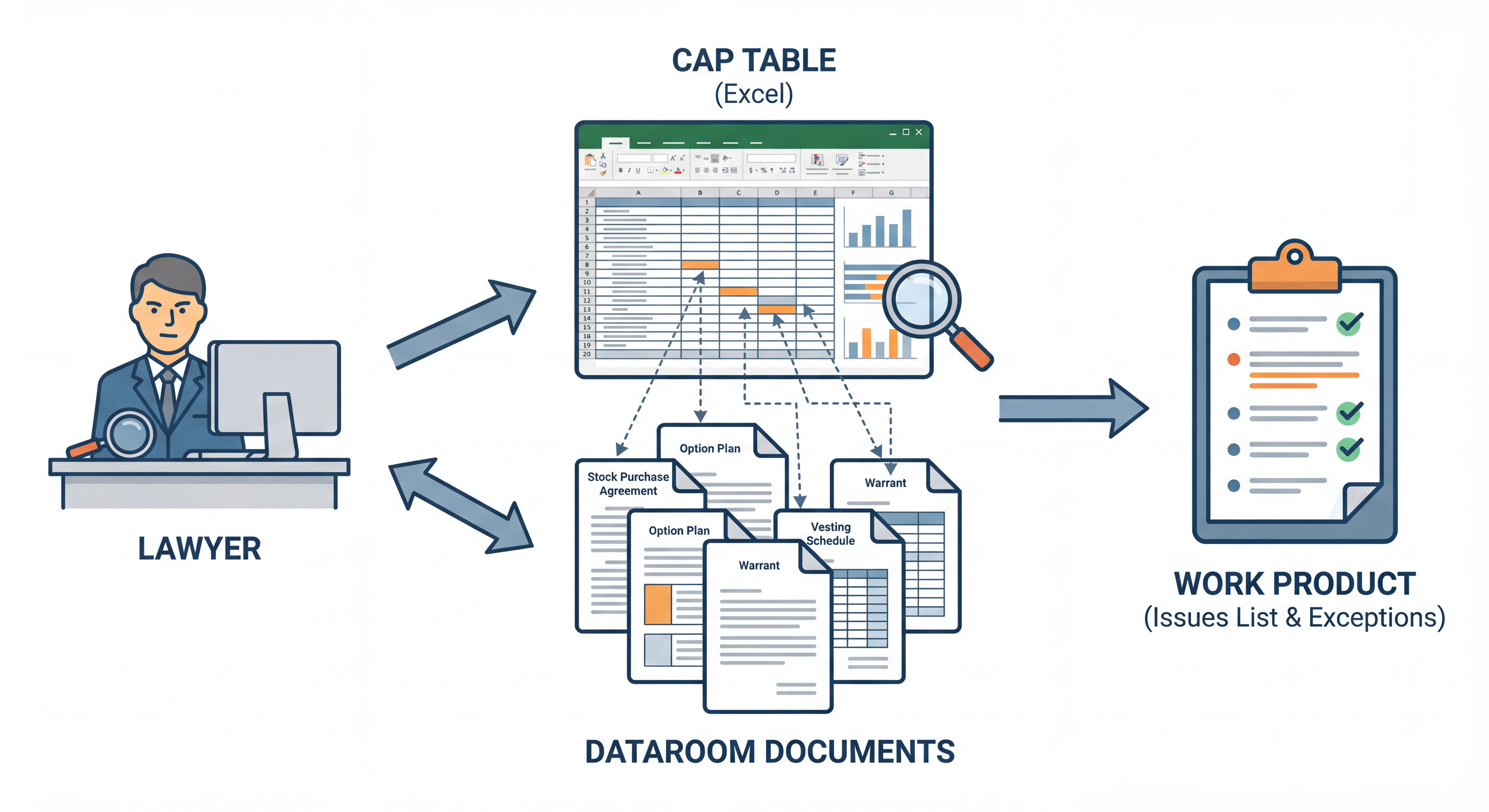}
    \vspace{-.1cm}
    \caption{
        \emph{The tie-out workflow}. Lawyers review  and cross-reference heterogeneous dataroom documents—one or more capitalization tables, and supporting legal documentation (e.g., SAFEs, option agreements, amendments, cancellations, etc.)—to understand the legal reality of the company and ensure the capitalization table is accurate by comparing it against the non-cap table documents, which provide the ground truth. \emph{The output}: verified legal positions 
        with document traceability, and flags for discrepancies requiring further review and/or action. 
    }
    \label{fig:tieout-workflow}
\end{figure}

\textbf{Tie-out} is the core verification task to reconcile legal ownership and related positions. Each entry in the capitalization table, such as share issuances, option grants, and SAFE conversions must trace to signed agreements, board approvals, and payment records. As shown in Figure~\ref{fig:tieout-workflow}, tie-out transforms a heterogeneous document collection into verified ownership records while surfacing discrepancies for legal review. The task is combinatorial: a single stock grant may depend on the original equity plan, a board consent, the signed option agreement, and subsequent amendments. Errors propagate, and even a single missed cancelation or overlooked amendment can misstate ownership across dozens of stakeholders.

\textbf{Why LLMs alone are not enough.} Tie-out exposes limitations that legal AI benchmarks do not test. Unlike single-document question answering \cite{tuggener2020ledgar,chalkidis2019neural,guha2022legalbench} or clause extraction \cite{guha2023legalbench}, tie-out requires \emph{multi-document reasoning} combining information scattered across dozens of files. It demands \emph{strict traceability} in that every claim must link to source evidence. And it requires \emph{consistency}: the same document processed twice must yield identical outputs. These requirements challenge the dominant agentic LLM paradigm, where performance excels at fluent extraction and summarization but struggles with combinatorial verification and reproducibility at scale \cite{cemri2025multi}.

This paper examines how LLMs can be effectively applied to capitalization tie-out, characterizing the complexity of the task and the architectural requirements for its reliable automation.
\begin{figure}
  \centering
  \begin{minipage}{0.45\textwidth}
    \centering
    \includegraphics[width=\textwidth]{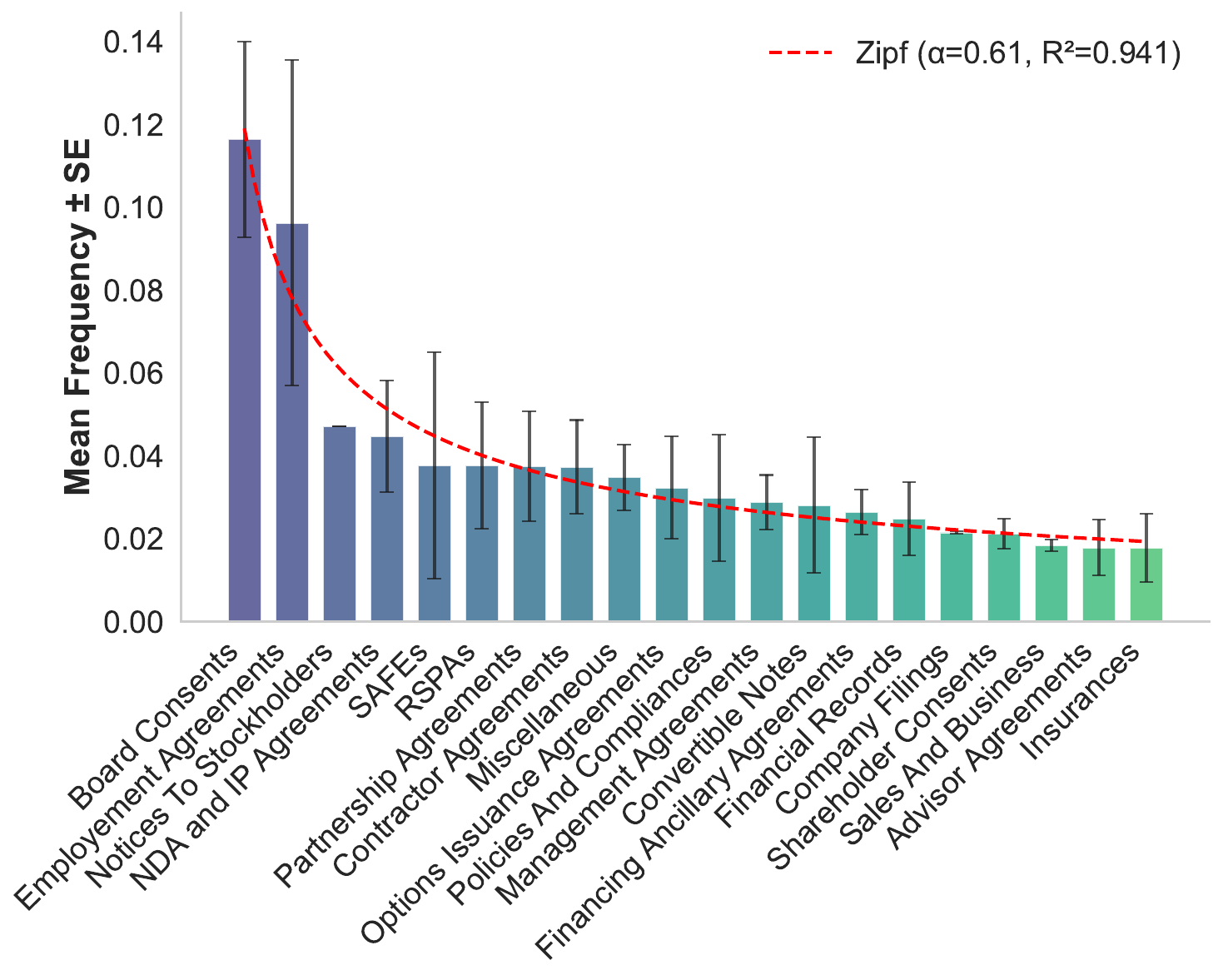}
  \end{minipage}
  \hfill
  \begin{minipage}{0.45\textwidth}
    \centering
    \includegraphics[width=\textwidth]{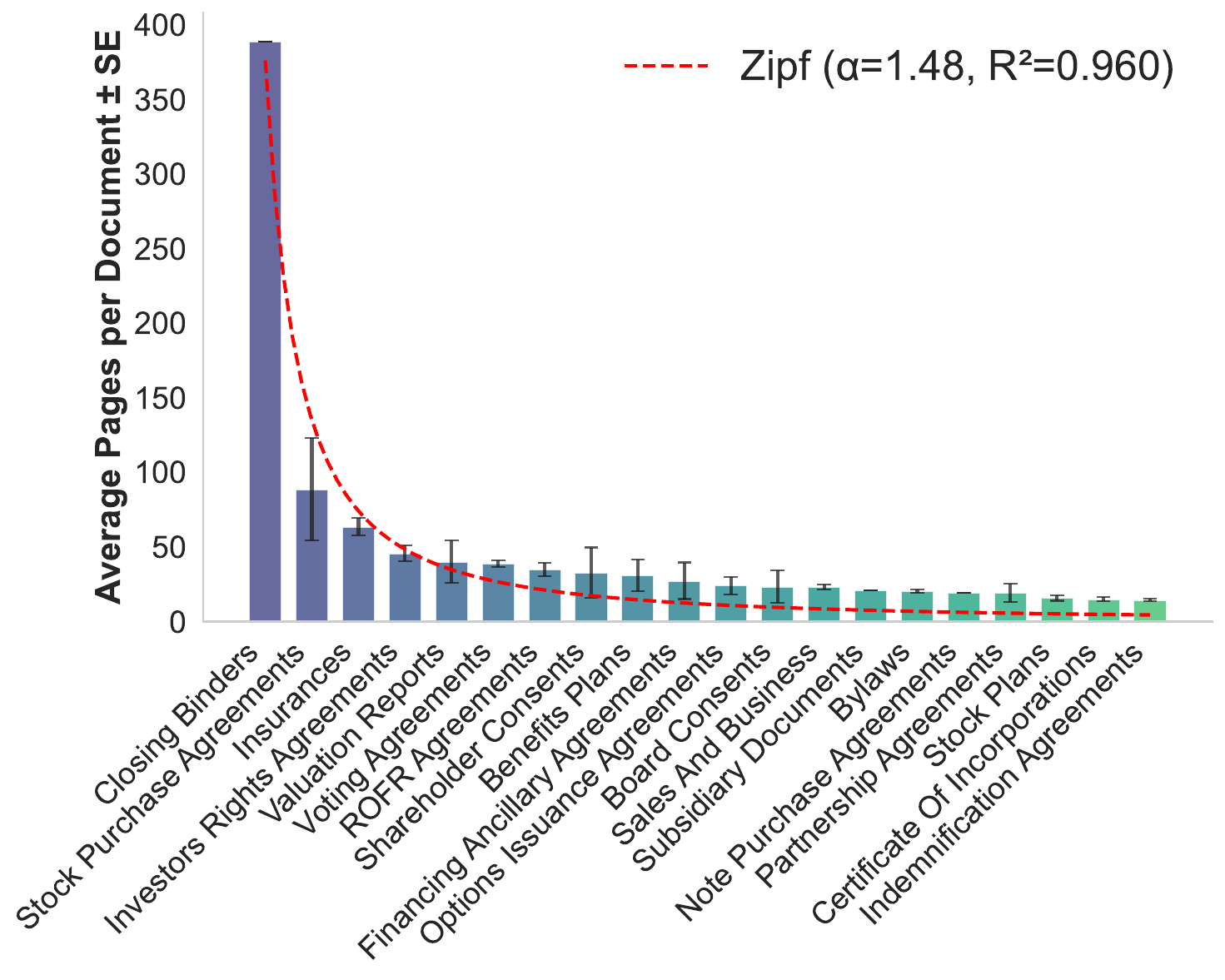}
  \end{minipage}
\caption{Document category. Red dashed lines indicate Zipf's law fits ($R^2 \sim 1$ is perfect fit).}
    \vspace{-.5cm}
\end{figure}

\section{Background and Formulation}\label{sec:problem}

    This section describes the capitalization due diligence context, introduces the dataroom and its core artifact—the cap table—and formally defines the tie-out problem.
    \subsection{The Dataroom}\label{subsection:dataroom}
    In any financing or M\&A transaction, the \emph{dataroom} is a repository containing the company's legal and financial history. We represent it as $\mathcal{D} = \{D_1, \dots, D_N\}$, where each document $D_i$ encodes partial, overlapping, or sometimes conflicting information about ownership, rights, and past transactions.
    
    Documents vary widely in formats, from contracts, registers, notices, certificates to spreadsheets. But each belongs to a functional category with a distinct role in verification (Table~\ref{tab:dataroom-docs}). As shown in Figure~\ref{fig:doc-composition}, the distribution follows Zipf's law: a small number of categories (e.g. board consents, SPA) account for the majority of documents, while a long tail of specialized instruments appears infrequently but remains critical when present \cite{piantadosi2014zipf,newman2005power}.

    \subsection{The Capitalization Table}\label{subsection:cap_table}
        The \emph{capitalization table} (cap table), denoted $\captable$, plays a central role in the tie-out. It provides a consolidated snapshot of all outstanding securities—common stock, preferred shares, options, SAFEs—along with their holders and ownership percentages. It typically takes the form of an Excel spreadsheet, with each tab (or \textit{ledger}) detailing the current ownership for one specific class of shares. An example of such \textit{ledger} is provided in Appendix \ref{appendix:ledger_example}. 
        
        In the tie-out problem, the ledgers from the cap table functions as the \emph{master claims ledger}: every entry must trace to authoritative source documents in the dataroom. The goal is not to reconstruct the company's full legal history but to confirm that each line is supported by controlling evidence—and to flag where it is not.
        
        \begin{table}
        \centering
        \caption{Document categories for capitalization tie-out in venture financings}
        \label{tab:dataroom-docs}
        \resizebox{0.75\textwidth}{!}{%
        \begin{tabular}{p{5cm} p{9cm}}
        \toprule
        \textbf{Document Type} & \textbf{Role in Tie-Out} \\
        \midrule
        Capitalization Table (Cap Table) & Records ownership of Company stock, convertible securities, and purchase rights; used as the purported "correct" representation of the Company's ownership against which non-cap table documents are compared. \\
        Stock Purchase Agreements & Document purchases of Company stock by stockholders; used to verify stock ownership and contractual rights. \\
        Convertible Securities & Document purchases of convertible securities (e.g., SAFEs, convertible notes) by investors; used to verify convertible securities ownership and conversion rights. \\
        Employment Contracts & Document equity ownership promised by the Company to employees, contractors, and advisors; used to verify that there are no promised equity unrepresented on the cap table. \\
        Amendments & Modify prior agreement terms (e.g., vesting schedule, acceleration terms); used to reconcile historical changes in ownership and document terms. \\
        Certificates of Incorporation / Bylaws & Establish the Company's legal existence, share class structure, and governance rules; used to verify issued securities comply with legal limit of authorized shares. \\
        Board and Shareholder Consents & Approve corporate actions (e.g., stock issuances, amendments, or corporate actions); used to validate formal board and/or stockholder authorization of transactions. \\
        Term Sheets & Outline terms of proposed financing; used to cross-check final agreements and conversion terms. \\
        Ancillary Equity Agreements (e.g., Option Exercise, Repurchase, and Transfer) & Document additional equity-related events; used to ensure timely and accurate  reflection of outstanding ownership on the cap table. \\
        \bottomrule
        \end{tabular}%
        }
    \end{table}
    
    \subsection{The Tie-Out Problem}\label{subsection:tieout_problem}
    
    At a high level, tie-out is a reconciliation problem. The company presents a \emph{reference} (or \emph{nominal}) cap table, while the dataroom—charters, board consents, SAFEs, warrants, option plans, and so on—provides the ground truth legal materials, implicitly defining a second, \emph{virtual} (or \emph{legal}) cap table. Tie-out asks whether these two capitalization states coincide, and if not, where and why they diverge.
    
    Formally, let $\mathcal{S} = \{S_1, \dots, S_M\}$ denote the set of securities (common stock, each preferred series, SAFEs, options, warrants, etc.), and let $\mathcal{D}$ denote the dataroom. We write
    \[
    \mathcal{X}
    \]
    for the space of possible capitalization states over $\mathcal{S}$ (authorized and issued amounts, liquidation preferences, conversion mechanics, option pools, and so on).
    
    The company’s cap table is a distinguished element
    \[
    \captableref \in \mathcal{X}
    \]
    (the \emph{reference} cap table). Conceptually, the dataroom induces a \emph{virtual} cap table
    \[
    \captablevirtual = T_{\mathrm{doc}}(\mathcal{D}) \in \mathcal{X},
    \]
    where $T_{\mathrm{doc}} : \mathcal{D} \to \mathcal{X}$ is the (typically implicit) extraction map that reads the legal documents and infers the capitalization they imply.
    
    Tie-out does not usually construct $\captablevirtual$ explicitly. Instead, it operates through a finite family of \emph{verification transforms}
    \[
    \mathcal{T} = \{T_1, \dots, T_K\},
    \qquad
    T_k : \mathcal{X} \to \mathcal{Y}_k.
    \]
    Each transform $T_k$, for each $k$, “zooms in” on a specific aspect of the cap table, for example "The date the board approved grant CS-001". Conceptually, the tie-out process checks whether the corresponding views of the virtual and reference cap tables coincide:
    \[
    T_k\bigl(\captablevirtual\bigr) \;\stackrel{?}{=}\; T_k\bigl(\captableref\bigr).
    \]
    The family of equalities
    \[
    T_k\bigl(\captablevirtual\bigr) = T_k\bigl(\captableref\bigr)
    \quad\text{for } k = 1,\dots,K
    \]
    are the \emph{tie-out constraints}. In practice, a lawyer may never write $\captablevirtual$ down; each verification task is just one instance of checking such an equality for a particular $T_k$ using the underlying documents.
    
    The outcome of tie-out is the set of \emph{anomalies} detected, i.e., the constraints that fail or cannot be established from the dataroom. We write
    \[
    \mathcal{A}
    =
    \big\{
    \,(k, T_k(\captablevirtual), T_k(\captableref), E_k)
    \;\big|\;
    T_k(\captablevirtual) \neq T_k(\captableref)
    \,
    \ E_k \subseteq \mathcal{D}
    \big\},
    \]
    where $E_k$ is the evidentiary subset of the dataroom used to compute (or fail to compute) the relevant quantities. The lawyer’s task is to produce $\mathcal{A}$ with full traceability: every recorded anomaly must be backed up by an explicit and minimal evidence set $E_k$.
    
    \paragraph{Coarse taxonomy of anomalies.}
    
    At this level of abstraction, different “types” of flags correspond to different ways in which tie-out constraints fail or cannot be established. Coarsely, we can distinguish:
    
    \begin{itemize}
      \item \textbf{Missing from cap table.} A position or security is present in the virtual cap table but absent from the reference cap table. In terms of the transforms, there exists a $k$ such that $T_k(\captablevirtual)$ encodes a nonzero position (e.g., issued shares, an outstanding warrant) while $T_k(\captableref)$ does not reflect it at all. Typical examples include a SAFE, warrant, or option grant that appears in the documents but is not carried onto the actual cap table.
    
      \item \textbf{Missing documentation.} A position or change is reflected on the reference cap table, but the dataroom does not fully determine the corresponding virtual view. Formally, $T_k(\captableref)$ is well-defined, while $T_k(\captablevirtual)$ is undefined or under-specified given $\mathcal{D}$ (e.g., a missing issuance, transfer, repurchase, or board consent). The anomaly is not a numerical mismatch but the absence or incompleteness of the evidentiary path that would support $T_k(\captableref)$.
    
      \item \textbf{Inconsistent terms.} Both cap tables carry the same “object” (e.g., a particular series, grant, or instrument), but the associated economic or administrative terms differ. In the transform view, $T_k$ isolates those terms (vesting schedule, acceleration provisions, price per share, liquidation preference, etc.), and the anomaly is simply
      \[
        T_k(\captablevirtual) \neq T_k(\captableref).
      \]
      Here the mismatch lives in the qualitative or parametric description of the security rather than its mere presence or absence.
    \end{itemize}

\begin{figure}
    \centering
    \includegraphics[width=\linewidth]{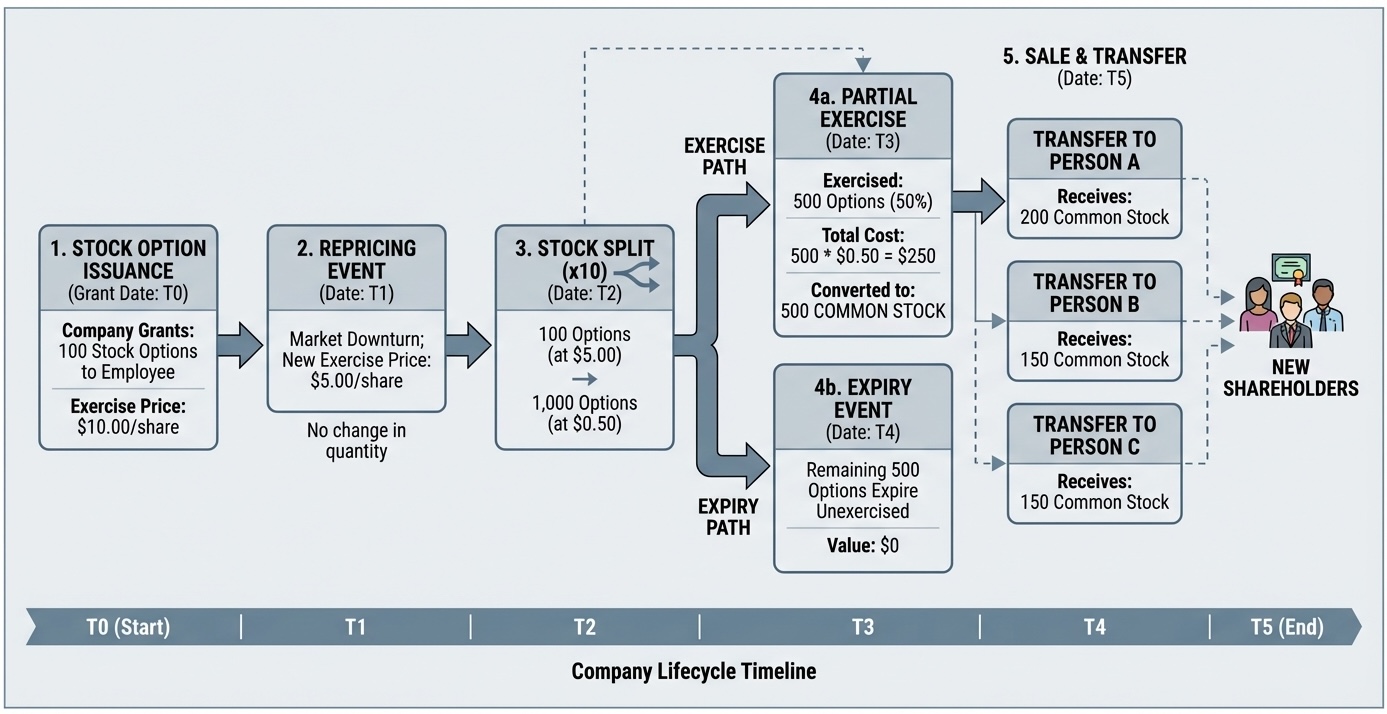}
    \caption{\emph{Real-world example of a grant lifecycle.} A single stock option grant that is later repriced, then affected by a 10:1 stock split, then partially exercised into common stock (with the remainder expiring), after which the resulting common shares are sold or transferred to three different holders.}
    \label{fig:grant-lifecycle}
    \end{figure}

    \begin{tcolorbox}[
      colback=black!5!white,
      colframe=black!75!white,
      title=\textbf{Challenges of Real-World Tie-Out at Scale},
      fonttitle=\bfseries
    ]
    \textbf{The lifecyle of grants.} The reference cap table provides only a partial view of the world: a compressed snapshot of nominal positions, stripped of the lifecycle that produced them. Each visible row aggregates a potentially long lineage of corporate actions—grants, repricings, stock splits, exercises, expirations, and transfers. As the company matures, those lineages branch and interact: verifying a single cap table row may require cross-referencing dozens of agreements spanning years of history. Figure~\ref{fig:grant-lifecycle} illustrates this on a concrete example. The combinatorics compound fast: 30 stakeholders, 5 security classes, and 200 documents already create thousands of distinct verification paths.
    
    \vspace{0.5em}
    \textbf{Fractured data landscape.} Documents are concatenated (one PDF containing an agreement, amendments, and exhibits), scanned with variable OCR quality, or missing pages. Near-duplicate versions proliferate: draft and executed copies, redlines, “clean” versions, and standalone signature packets, sometimes with only one of them actually operative. Filenames are misleading; metadata is absent. Temporal reasoning is difficult: amendments modify or supersede prior agreements, but the chain is rarely explicit. Documents come in separate batches and reference others not in the dataroom; parties appear under variant names or through affiliated entities.
    \end{tcolorbox}

\section{Empirical Complexity Analysis}\label{sec:empirical_complexity}

\begin{figure}[h]
    \centering
    \includegraphics[width=\linewidth]{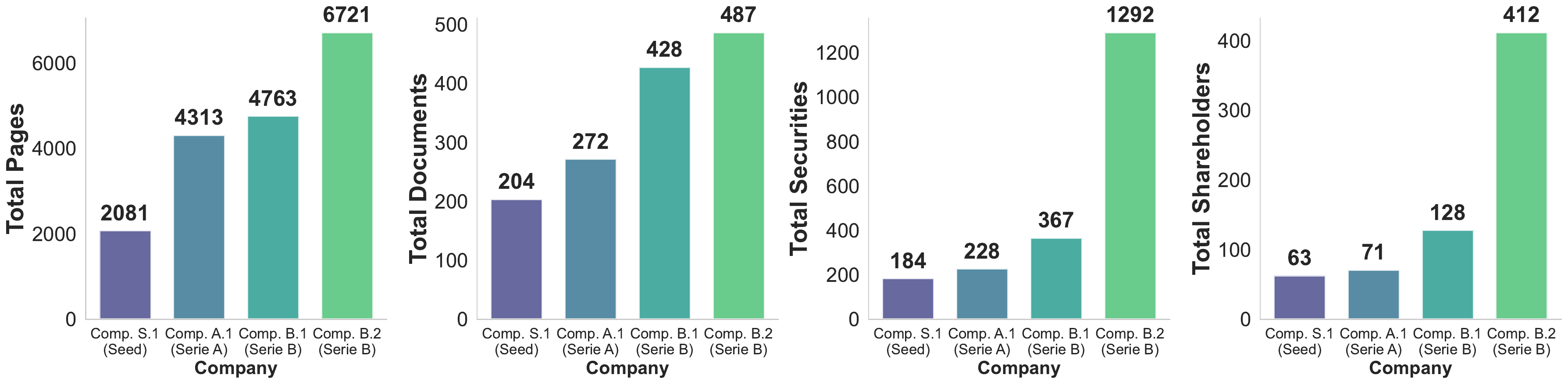}
    \caption{Comparison of key dataroom statistics, including total pages, documents, securities, and shareholders, across companies in different financing stages.}
    \label{fig:company-metrics}
\end{figure}

\begin{figure}[h]
    \centering
    \includegraphics[width=\linewidth]{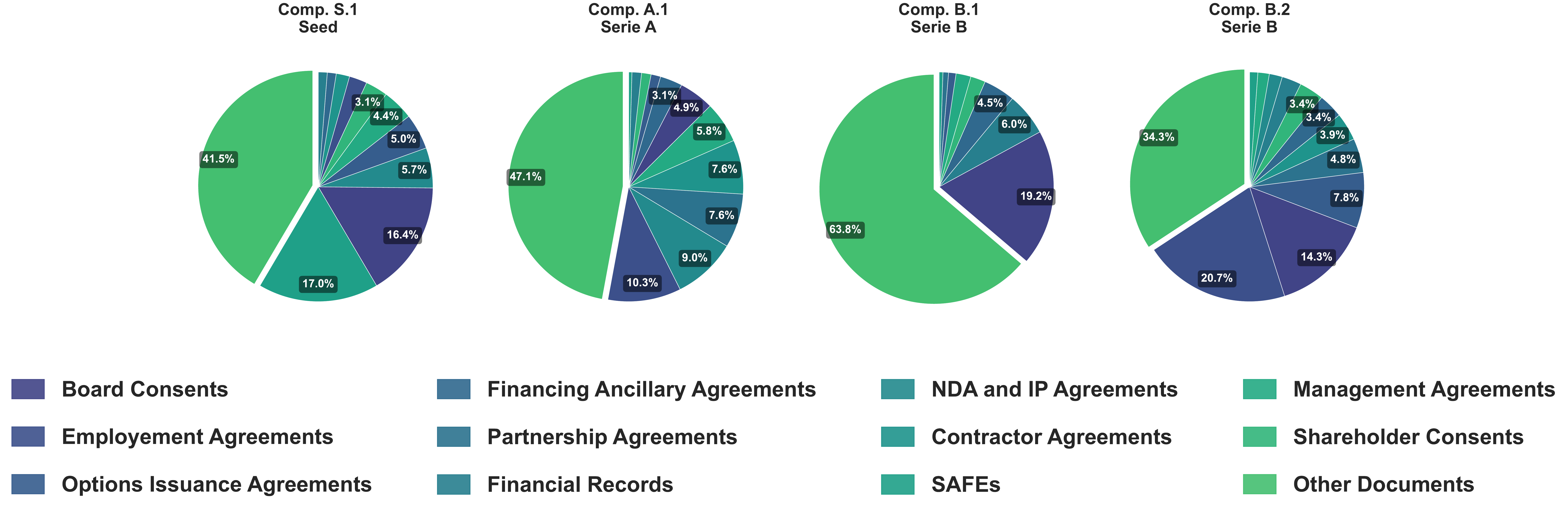}
    \caption{Distribution of categories across four companies at varying financing stages.}
    \label{fig:doc-composition}
\end{figure}

\begin{figure}[h]
    \centering
    \includegraphics[width=\linewidth]{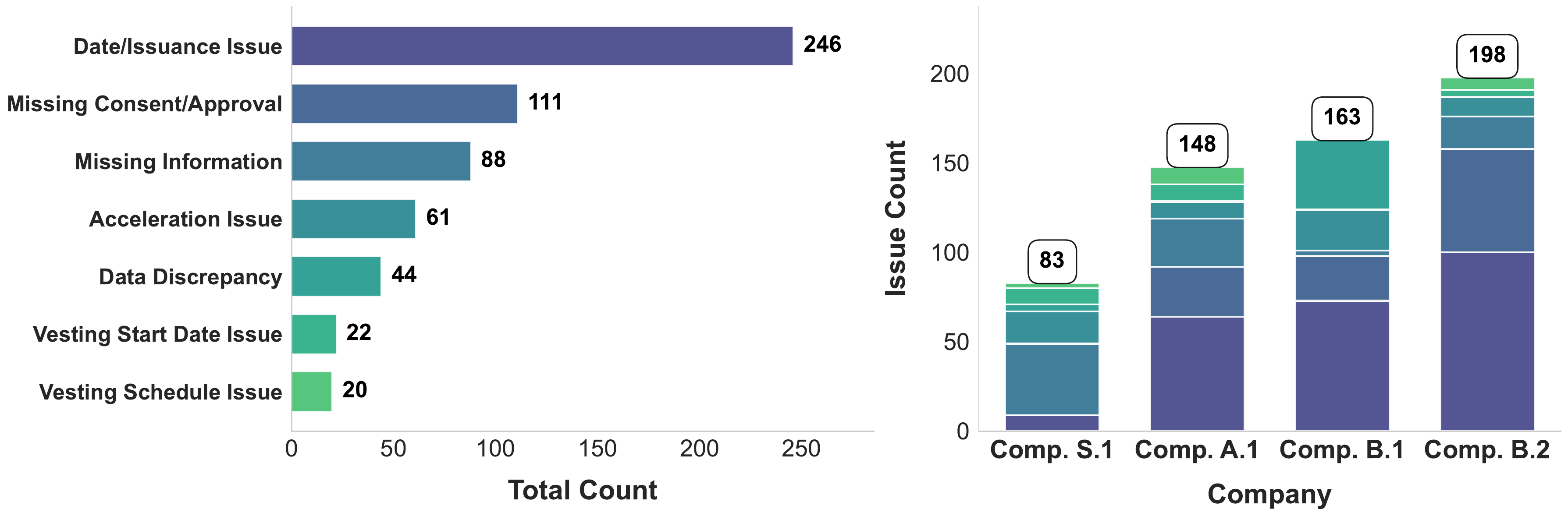}
    \caption{Breakdown of issues detected in real-world datarooms, showing the total frequency of each issue type (left) and their distribution across four different companies (right).}
    \label{fig:issues-distribution}
\end{figure}

\begin{wrapfigure}{l}{0.42\linewidth}
    \centering
    \vspace{-10pt} 
    \includegraphics[width=\linewidth]{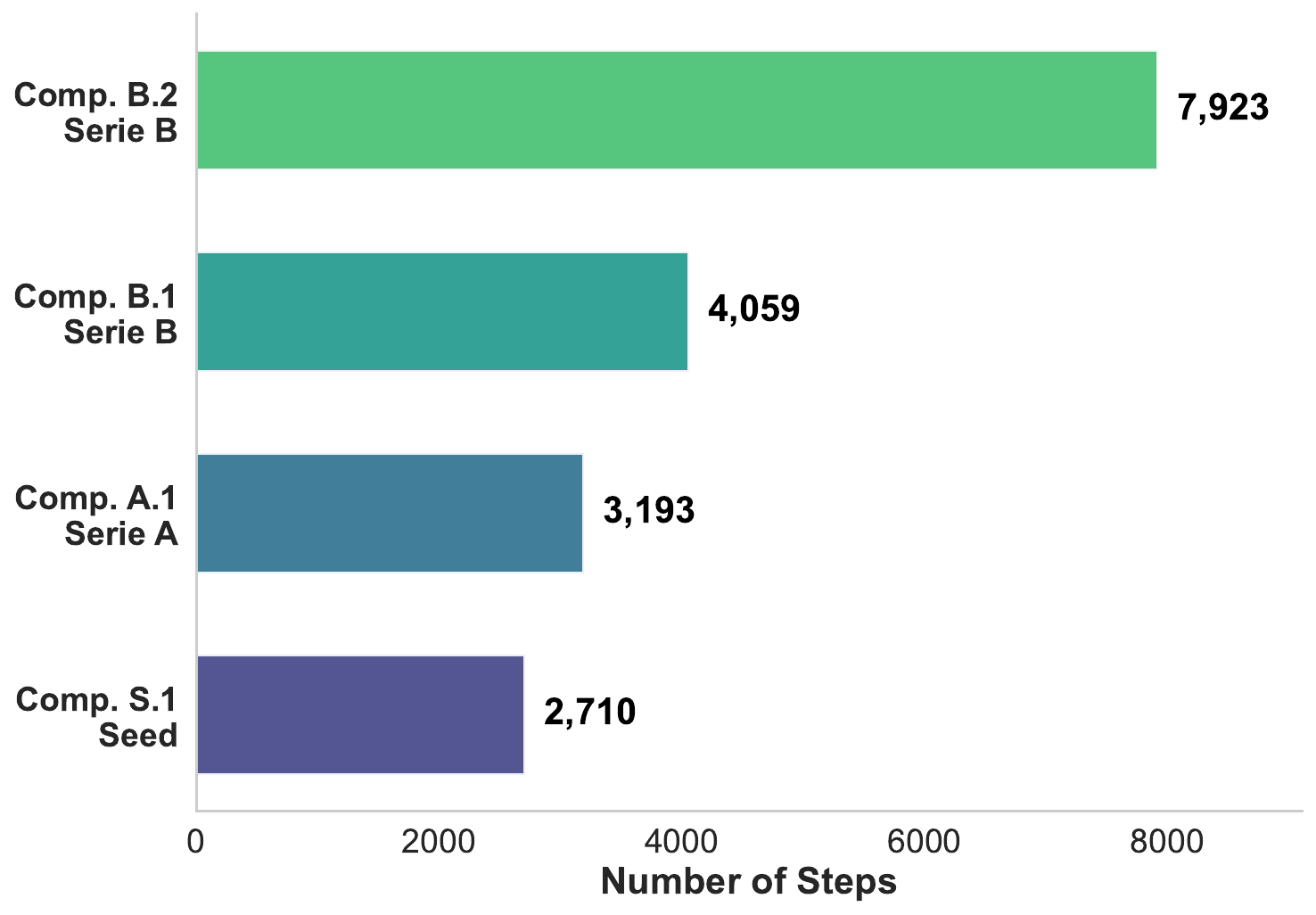}
    \caption{Number of verification steps.}
    \label{fig:steps_per_dataroom}
    \vspace{-10pt} 
\end{wrapfigure}

The formal definition above frames tie-out as mapping a dataroom $\mathcal{D}$ to a virtual cap table $\captablevirtual$ via extraction ($T_{\mathrm{doc}}$) and verification transforms ($T_k$). In an idealized setting, $\mathcal{D}$ is small, structured, and internally consistent. However, empirical data from real-world financing diligence reveals that the complexity of this process does not scale linearly. By analyzing dataroom statistics across four representative companies spanning Seed to Series B financing stages (Fig. \ref{fig:company-metrics}, Fig. \ref{fig:doc-composition}, Fig. \ref{fig:steps_per_dataroom}, and Fig. \ref{fig:issues-distribution}), we identify three primary drivers of complexity related to scaling evidence, shifting anomaly types, and verification workload.

\paragraph{The evidentiary burden scales super-linearly relative to document volume.}
Fig. \ref{fig:company-metrics} illustrates the velocity at which the diligence surface area expands. While the growth in raw volume—pages quadrupling and total documents more than doubling between Seed (Comp. S.1) and late Series B (Comp. B.2)—is significant, the critical insight lies in the changing relationship between documents and the securities they govern. While the size of the dataroom $|\mathcal{D}|$ roughly doubles, the number of individual securities tracked—the cardinality of the set $\mathcal{S}$—increases by a factor of seven (from 184 to 1,292). This widening gap indicates a fundamental shift in the required granularity of the extraction map $T_{\mathrm{doc}}$. At the Seed stage, the ratio of documents to securities is roughly 1:1. By Series B, a single document in $\mathcal{D}$ (such as a major recapitalization agreement) may define hundreds of distinct security issuances. Consequently, the verification transforms $T_k$ must become increasingly granular to isolate individual data points buried within dense legal text.

\paragraph{As governance matures, anomalies shift from informal omissions to complex inconsistencies.}
As the volume expands, the qualitative nature of the "source of truth"—and the errors it contains—evolves. Fig. \ref{fig:doc-composition} shows that Seed stage datarooms are dominated by unstructured documents like employment agreements, reflecting early-stage informal governance. Here, lawyers must scan unstructured text to identify where business intent never translated into formal evidence, leading to rudimentary "Missing Information" anomalies. By Series B, governance is structured, dominated by dense financing records. The challenge shifts to "catching up" on years of intricate historical transactions to spot inconsistencies between interrelated documents. Fig. \ref{fig:issues-distribution} confirms that total anomaly counts grow significantly with maturity—Comp. B.2 surfaces nearly $2.5\times$ the issues of Comp. S.1. Notably, the persistence of "Missing Information" and "Missing Consent/Approval" as top categories in later stages indicates that the primary challenge in mature tie-outs becomes verifying the *completeness* of an exponentially growing historical record—a "needle in a haystack" retrieval problem that human reviewers struggle to scale.

\paragraph{The verification workload explodes in both volume and cognitive difficulty.}

The combined effects of super-linear evidence growth and increasingly buried anomalies materialize as a severe increase in the verification workload of the lawyer. Fig. \ref{fig:steps_per_dataroom} quantifies this burden by tracking the total number of atomic "steps" executed by the counsel to complete the tie-out, acting as a practical proxy for $K$, the total family of required verification transforms. We observe a near-tripling of workload, from approximately 2,700 steps at Seed to nearly 8,000 steps at Series B. Critically, this increase is not merely numerical; each step becomes intrinsically harder to perform. As the haystack ($\mathcal{D}$) grows, the retrieval task required for every individual check becomes more cognitively demanding for a human reviewer. In practice, the tie-out process often turns into a grueling exercise over tens of hours of burdensome review.
    
\section{Towards Tie-Out Automation}
\label{sec:towards_automation}

    The core technical challenge of tie-out is bridging the semantic gap between highly unstructured, heterogeneous legal documents ($\mathcal{D}$) and the rigid, combinatorial constraints required for verification ($T_k(\mathcal{C}^{\text{virt}}) \stackrel{?}{=} T_k(\mathcal{C}^{\text{ref}})$).
    
    We investigate two fundamentally different architectural paradigms for addressing this challenge: a "lazy" approach that attempts extraction and verification simultaneously via agentic reasoning, and an "eager" approach (Equall) that decouples the process by explicitly constructing a layered, structured world model.
    
    \subsection{The Agentic Paradigm: Lazy Construction via RAG}\label{subsection:agentic_paradigm}
    
    The dominant paradigm for tackling complex, document-based tasks with LLMs is the \textit{agentic framework} augmented by Retrieval-Augmented Generation (RAG). In this approach, construction of the virtual view $\mathcal{C}^{\text{virt}}$ is "lazy"—it occurs ad-hoc in response to specific verification queries.
    
    \textbf{How it works:} To verify a specific transform $T_k$ (e.g., "Verify vesting start date for Option Grant CS-102"), an LLM agent orchestrates a multi-step process:
    \begin{enumerate}[nosep]
        \item \textbf{Query Generation:} The agent formulates search queries using the target field and entities.
        \item \textbf{Retrieval:} A retrieval system fetches relevant chunks from the indexed dataroom $\mathcal{D}$.
        \item \textbf{Reasoning \& Extraction:} The LLM synthesizes retrieved chunks to extract the requested value and identify supporting evidence $E_k$.
        \item \textbf{Verification:} The agent compares the extracted value against the reference cap table $\mathcal{C}^{\text{ref}}$.
    \end{enumerate}
    
    \textbf{Limitations in the Due Diligence Context:} While flexible, this paradigm faces severe challenges at the scale and complexity described in Section \ref{sec:empirical_complexity}. First, standard RAG struggles with \textbf{"global" reasoning constraints}. Proving a "Missing Documentation" anomaly requires establishing that evidence does \textit{not} exist across thousands of pages—a task where retrieval failure is often indistinguishable from actual absence. Second, \textbf{compounding errors in lineage tracking}. Verifying a single position may require tracing a multi-year chain of dependent documents; the probability of an agent successfully retrieving and correctly reasoning over every link in that chain decreases exponentially with chain length.
    
    \subsection{The Equall Paradigm: Eager Construction of a World Model}\label{subsection:equall_paradigm}
    
    To overcome the limitations of ad-hoc reasoning, we propose a fundamentally different approach: decoupling extraction from verification through the eager construction of a \textit{symbolic world model}. This model is built in two stages, moving from raw text to atomic facts, and finally to a strong inductive representation of the company's lifecycle.
    
    \paragraph{Stage 1: Foundational Extraction (Low-Level Nodes).}
    The first stage processes the raw dataroom $\mathcal{D}$ to build the foundation of a knowledge graph. We employ specialized LLM-based parsers to first classify documents according to the taxonomy established in Fig. 2. Subsequently, extracting agents identify and instantiate low-level nodes such as \texttt{Stakeholders} (individuals, funds), \texttt{Securities} (specific stock classes, warrants), and structured atomic values (dates, share counts, prices, clauses), linking each explicitly to its source document span for provenance.
    
    \paragraph{Stage 2: Inductive Event Modeling (Conceptual Nodes).}
    The critical innovation lies in the second stage: organizing these low-level facts into a coherent, temporal representation of the company's legal history. We define high-level "Conceptual Nodes" representing business events: \texttt{Issuance}, \texttt{Transfer}, \texttt{Amendment}, \texttt{Conversion}, \texttt{Exercise}, and \texttt{CorporateAction} (e.g., stock splits). LLM reasoning is used to synthesize low-level nodes into these event nodes. For example, an \texttt{Amendment} event node is constructed by linking the amending document, the specific clauses changed, and crucially, a relationship edge pointing to the prior \texttt{Issuance} or \texttt{Agreement} event it modifies. This stage results in a rich "Event Graph"—a strong inductive representation of the company's entire lifecycle. This structured history is a reusable asset whose utility extends beyond tie-out to adjacent legal tasks flowing from the same legal reality.
    
    \paragraph{Stage 3: Targeted Neuro-Symbolic Verification.}
    Finally, tie-out verification is executed via targeted queries over this Event Graph. We adopt a \textbf{neuro-symbolic} approach:
    \begin{enumerate}[nosep]
        \item The "neuro" component is the robust, LLM-driven extraction and event synthesis that built the graph (Stages 1 \& 2), handling the ambiguity of legal text.
        \item The "symbolic" component is the application of deterministic logic to aggregate these events into the final virtual cap table state $\mathcal{C}^{\text{virt}}$.
    \end{enumerate}
    
    For example, verifying a stakeholder's current share count is no longer a fuzzy retrieval task. It is a structured query: traverse the Event Graph for all \texttt{Issuance} events to that stakeholder, subtract subsequent \texttt{Transfer-Out} events, add \texttt{Transfer-In} events, and apply adjustments from linked \texttt{CorporateAction} events (like splits). The result of this targeted query is compared against $\mathcal{C}^{\text{ref}}$ to identify anomalies.
    

\section{Experiments}
    
    \paragraph{Task Definition.} We evaluate automated tie-out as the anomaly detection task defined in Section \ref{subsection:tieout_problem}. Given a dataroom $\mathcal{D}$ and a reference capitalization table $\mathcal{C}^{\text{ref}}$, the system must produce the set of anomalies $\mathcal{A}$. For every verification transform $T_k$ where the virtual view differs from the reference view ($T_k(\mathcal{C}^{\text{virt}}) \neq T_k(\mathcal{C}^{\text{ref}})$ or is undefined), the system must flag the discrepancy, classify its type, identify the affected stakeholder/security, and provide the supporting evidentiary subset $E_k \subseteq \mathcal{D}$.
    
    \paragraph{Harmonized Flag Types.} To connect empirical evaluation with our theoretical framework, we categorize anomalies based on the taxonomy established in Section \ref{subsection:tieout_problem}:
    \begin{itemize}[nosep]
        \item \textbf{Terms Discrepancy ($T_k(\mathcal{C}^{\text{virt}}) \neq T_k(\mathcal{C}^{\text{ref}})$)}: The entry exists in both views but differs. Practically, this includes numerical mismatches (share counts, prices, dates) often labeled as ``Data Discrepancy'' or conflicting terms across agreements (``Issuance Discrepancy'').
        \item \textbf{Missing Documentation ($T_k(\mathcal{C}^{\text{virt}})$)}: An item on the reference cap table lacks sufficient supporting evidence in $\mathcal{D}$. Practical examples include ``Board Approval Missing'' for recorded issuances or broken chains of title.
        \item \textbf{Missing from Cap Table ($T_k(\mathcal{C}^{\text{ref}})$)}: Valid securities or stakeholders identified in $\mathcal{D}$ are absent from the reference cap table.
    \end{itemize}
    
    \paragraph{Dataset \& Metrics.} We evaluate on the four anonymized datarooms presented in Section \ref{sec:empirical_complexity}, spanning Seed to Series~B. Ground-truth flags were annotated by experienced legal professionals. We report precision, recall, and F1 per flag category; a prediction is correct only if the type of anomaly and supporting evidence match the ground truth.

    \paragraph{Baselines.} We evaluate the paradigms detailed in Section \ref{sec:towards_automation} by comparing three approaches to constructing and verifying the virtual view $\mathcal{C}^{\text{virt}}$:
    \begin{itemize}[nosep]
        \item \textbf{Agentic Baseline}: Represents the "lazy construction" paradigm (Section \ref{subsection:agentic_paradigm}). It uses GPT-5.1 with iterative RAG and multi-step reasoning to perform extraction and verification ad-hoc for each query directly from raw documents.
    
        \item \textbf{Agentic + Structured Repr.}: An ablation operating over pre-extracted low-level nodes (Stage 2 of Equall's pipeline, see Section \ref{subsection:equall_paradigm}) rather than raw text. It relies on the ad-hoc, "lazy" agentic reasoning pipeline to dynamically connect evidence during verification.
    
        \item \textbf{Equall (Ours)}: The full "eager construction" approach (Section \ref{subsection:equall_paradigm}). It first builds the complete, layered world model—progressing from low-level extraction to the inductive Event Graph representing the company's lifecycle. Verification transforms $T_k$ are then executed as deterministic neuro-symbolic queries over this structured graph.
    \end{itemize}
    
    \subsection{Results}
\begin{figure}[h]
        \centering
        \includegraphics[width=\linewidth]{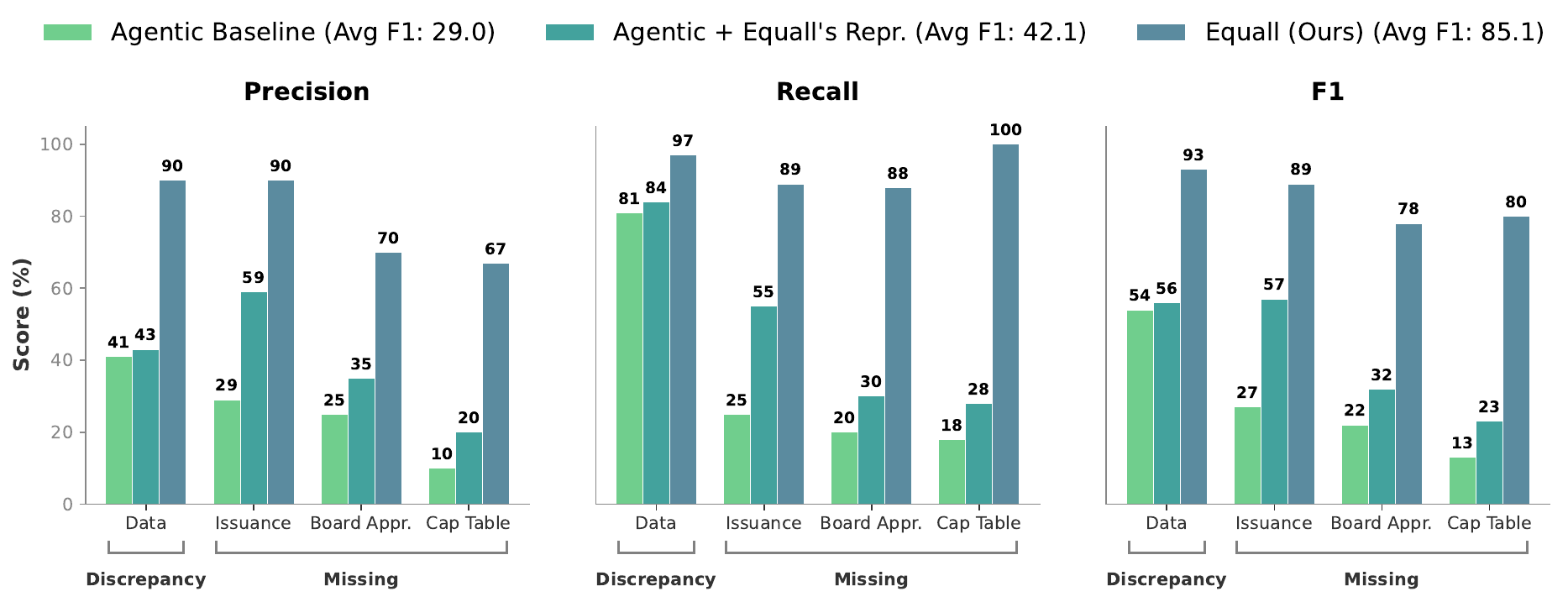}
        \caption{Precision, recall, and F1 across harmonized flag categories.}
        \label{fig:benchmark-results}
    \end{figure}

    \paragraph{Overall Performance.} Figure~\ref{fig:benchmark-results} shows results across flag types. Equall achieves an average F1 of 85\%, significantly outperforming agentic + structured representations (42\%) and pure agentic (29\%).
     The performance gap is interpreted through our theoretical lens. Agentic baselines perform respectably on \textbf{Inconsistent Terms} (e.g., Data Discrepancy), which are often "local" comparison tasks requiring the retrieval of only one or two source documents. However, agentic performance collapses on \textbf{Missing Documentation} and \textbf{Missing from Cap Table}. These anomalies require "global" reasoning over the dataroom (i.e., the legal source of truth) to prove a negative or establish a complete lineage. As analyzed in Section \ref{sec:empirical_complexity}, the non-linear scaling of evidentiary volume makes constructing this global context via ad-hoc retrieval statistically precarious. Equall's pre-computed knowledge graph turns these complex reasoning chains into reliable graph queries, providing consistent gains across the board.

    \paragraph{The Speed Trade-off: Lazy vs. Eager Construction.}
        This comparison highlights the fundamental architectural trade-off between the paradigms detailed in Section \ref{sec:towards_automation}. The agentic approach relies on "lazy," ad-hoc reasoning at query time. In contrast, Equall invests in the "eager" construction of the layered world model—transforming raw text into the inductive Event Graph before the first check is run. This shifts the computational burden of reasoning out of the critical verification path.
        
        \begin{figure}[h]
        \centering
        \begin{minipage}[t]{0.45\textwidth}
          \vspace{0pt}
          \centering
          \small
          \begin{tabular}{lcc}
          \toprule
          & \textbf{Agentic} & \textbf{Equall} \\
          \midrule
          \multicolumn{3}{l}{\textit{Indexing (one-time)}} \\
          \quad Time & 2 mins & 15 mins \\
          \midrule
          \multicolumn{3}{l}{\textit{Inference (per check)}} \\
          \quad Time & 45 sec & 2 sec \\
          \midrule
          \textit{100 checks} & 77 mins & 18 mins \\
          \textit{500 checks} & 377 mins & 32 mins \\
          \bottomrule
          \end{tabular}
        \end{minipage}
        \hfill
        \begin{minipage}[t]{0.50\textwidth}
          \vspace{0pt}
          \small
          The agentic baseline embodies the "lazy" paradigm: minimal setup cost, but a high marginal cost per verification due to repeated, complex ad-hoc reasoning. Equall's "eager" approach requires upfront investment to construct the Event Graph. However, this transforms verification into inexpensive, deterministic graph traversals (the "symbolic" phase of our approach). Equall's approach offers a critical 22$\times$ speed advantage per check, essential for responsive human-in-the-loop workflows.
        \end{minipage}
        \caption{Speed comparison on a 300-document dataroom. Eager world model construction amortizes reasoning costs over subsequent verification steps.}
        \label{tab:latency}
        \end{figure}

\begin{figure}[h]
    \centering
    \begin{minipage}[t]{0.48\textwidth}
    \vspace{0pt}
    \centering
    \includegraphics[width=\textwidth]{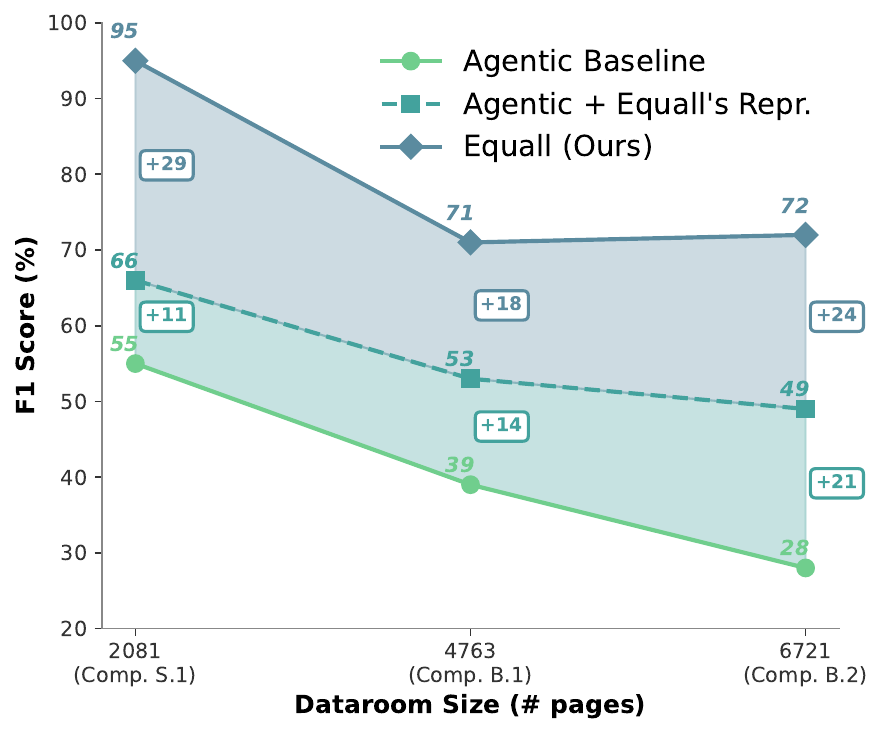}
    \caption{\textbf{Scaling behavior.} F1 on flag detection vs.\ dataroom size. Agentic quality degrades rapidly with complexity, while Equall system remains robust even on large series-B datarooms.}
    \label{fig:scaling}
    \end{minipage}
    \hfill
    \begin{minipage}[t]{0.48\textwidth}
    \vspace{0pt}
    \centering
    \includegraphics[width=\textwidth]{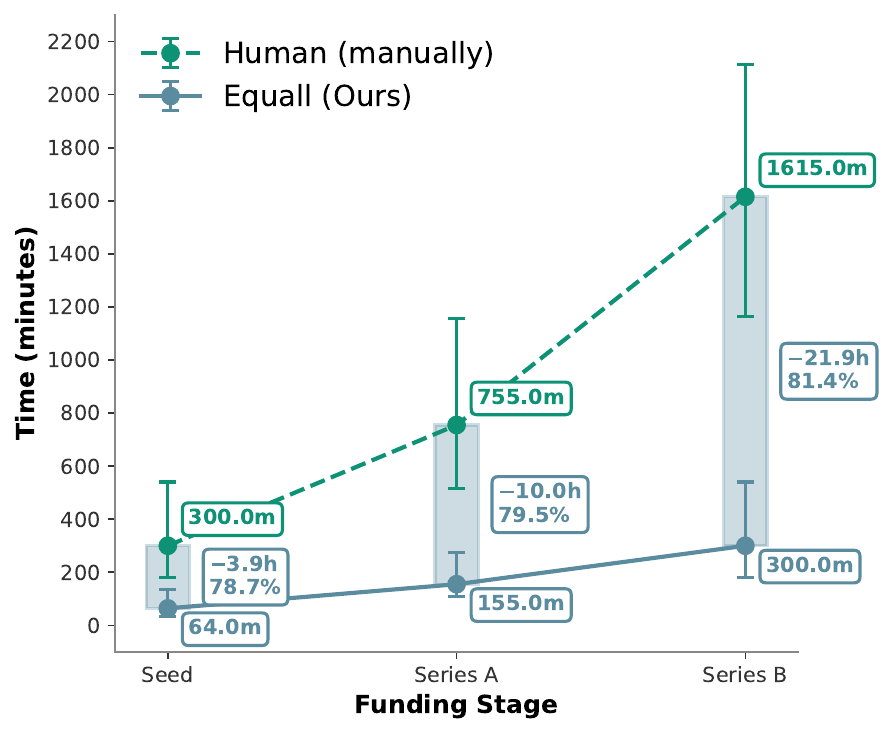}
    \caption{\textbf{Tie-out time: manual vs. Equall-assisted.} Error bars represent 95\% intervals obtained by comparing the time reported by customers/partners on real-world tieouts vs. end-to-end time obtained internally by using Equall’s system.}

    \label{fig:time-savings}
    \end{minipage}
    \end{figure}

    \paragraph{Real-World Efficiency at Scale.} Our experiments empirically validate the non-linear scaling of diligence complexity discussed in Section \ref{sec:empirical_complexity}. As datarooms grow in volume and semantic density (from Seed to Series B), the verification burden explodes for both automated baselines and human professionals. Figure~\ref{fig:scaling} shows the agentic baseline buckling under retrieval noise and compounding errors, with F1 scores dropping sharply from 55\% to 28\%. Conversely, Equall’s structured, "eager" modeling remains robust, widening the performance gap significantly at later stages. This complexity crisis is mirrored in manual workflows: Figure~\ref{fig:time-savings} reveals that human effort scales super-linearly, growing more than fivefold from 5 hours at Seed to nearly 27 hours at Series B. Equall effectively breaks down this complexity through upfront world modeling, resulting in a smooth scaling curve for assisted review (64m $\to$ 300m). By combining high-recall automated verification with targeted, high-precision human review, the Equall-assisted workflow delivers massive efficiency gains—scaling from roughly 79\% at Seed to 81.5\% at Series B—precisely where manual efforts become unsustainable.
    
    \paragraph{World Model as Multi-Purpose Utility.} Equall's massive improvement over the \texttt{Agentic + Structured Repr.} baseline shown in Figure~\ref{fig:benchmark-results} highlights the architectural importance of the intermediate "Event Graph" (Stage 2 in Section \ref{sec:towards_automation}) in driving performance. Notably, this inductive representation models generic corporate lifecycle events—issuances, amendments, transfers—rather than task-specific tie-out logic. The fact that a strongly typed, yet task-agnostic, structure yields state-of-the-art results on a highly specialized verification task suggests that the Event Graph successfully captures the fundamental legal reality of the company. This indicates that Equall's world model is not merely a single-purpose utility, but a robust foundational substrate suitable for a wider array of downstream legal applications that rely on the same underlying historical ground truth.
    
\section{Conclusion: A Framework for Applied Legal Intelligence}

    \paragraph{Towards Autonomous Tie-Out Agents.}
    The cap-table tie-out problem represents the critical frontier for deploying truly autonomous systems in high-stakes legal scenarios. Realizing this potential requires converging three essential ingredients, all present in this domain. First, a dense, verifiable reward signal: unlike subjective legal tasks, tie-out provides objective ground truth for training robust policy networks \cite{shao2024deepseekmath}. Second, scalable training environments: our empirical anomaly taxonomy enables the algorithmic generation of vast synthetic curricula by injecting known error patterns into validated "clean" datarooms. Third, and most critically for handling massive unstructured state spaces, a robust world model. Standard agents struggle with ad-hoc retrieval over raw text. The layered "world model" developed in Equall’s approach—specifically the inductive Event Graph described in Section \ref{subsection:equall_paradigm}—provides the necessary structured, temporal memory bank. Integrating this foundational world model with RL-driven training on synthetic data will enable the next generation of agents to navigate complex, multi-step lineage tasks with superhuman reliability, fundamentally transforming the practice of capitalization due diligence and comparable legal risk analysis work.

    \paragraph{Generalized Representation as a Foundation for Legal Intelligence.}
Equall's superior performance owes substantially to its explicit, inductive world model rather than to task-specific engineering. The Event Graph captures legally operative events as structured, temporally ordered state transitions grounded in primary evidence. Because these event primitives recur across legal domains, the architecture is inherently generalizable. Verification over this model reduces to deterministic queries on a well-defined state space, yielding stable performance even under substantial combinatorial complexity. Together, these findings suggest that explicit world-model construction is a promising architectural foundation for building reliable autonomous systems for applied legal reasoning.

\newpage
\bibliographystyle{plain}
\bibliography{equall}

\appendix

\include{ledger}

\end{document}

%% file: ledger.tex
\begin{landscape}

\section{Example of Common Stock Ledger}\label{appendix:ledger_example}

\begin{longtable}{@{}l l r l r l l@{}}
    \caption{A trimmed example of a typical Common Stock (the most common class of shares issued) ledger. Every Common Stock grant that was ever issued by the company should appear on this ledger in its most up-to-date state (i.e reflecting any subsequent amendments or corporate events that could affect it). We call the full capitalization table $\captableref$ the ensemble of all the ledgers submitted for the diligence process.} \label{tab:filtered_captable} \\
    \toprule
    \textbf{Security ID} & \textbf{Stakeholder Name} & \textbf{Quantity Issued} & \textbf{Share Class} & \textbf{Price Per Share} & \textbf{Vesting Schedule} & \textbf{Acceleration} \\
    \midrule
    \endfirsthead
    
    \toprule
    \textbf{Sec ID} & \textbf{Name} & \textbf{Qty Issued} & \textbf{Class} & \textbf{Price} & \textbf{Vesting} & \textbf{Accel.} \\
    \midrule
    \endhead
    
    \midrule
    \multicolumn{7}{r}{\textit{Continued on next page...}} \\
    \endfoot
    
    \bottomrule
    \endlastfoot
    
    CS-01 & Paul Reynolds & 3,162,500 & Common (CS) & \$0.00001 & 1/60th monthly, 1 year cliff & Double Trigger \\
    CS-02 & Sarah Lawson & 700,000 & Common (CS) & \$0.00001 & 1/60th monthly, 1 year cliff & Double Trigger \\
    CS-03 & Thomas Alvarez & 262,500 & Common (CS) & \$0.00001 & 1/60th monthly, 1 year cliff & Double Trigger \\
    CS-04 & Julien Moreau & 262,500 & Common (CS) & \$0.00001 & 1/60th monthly, 1 year cliff & Double Trigger \\
    CS-05 & Zara Ryman & 200,000 & Common (CS) & \$0.00001 & 1/60th monthly, 1 year cliff & Double Trigger \\
    CS-06 & Leigh Bartlett & 150,000 & Common (CS) & \$0.00001 & 1/60th monthly, 1 year cliff & Double Trigger \\
    CS-07 & Tim Branson & 150,000 & Common (CS) & \$0.00001 & 1/60th monthly, 1 year cliff & Double Trigger \\
    CS-08 & Lucas Costa & 50,000 & Common (CS) & \$0.00001 & 1/30th monthly, no cliff & Double Trigger \\
    CS-09 & David Velner & 37,500 & Common (CS) & \$0.00001 & 1/60th monthly, 1 year cliff & Double Trigger \\
    CS-10 & John Jackson & 5,000 & Common (CS) & \$0.00001 & 1/48th monthly, 1 year cliff & Yes \\
    CS-11 & Nadia Mansouri & 5,000 & Common (CS) & \$0.00001 & 1/48th monthly, 1 year cliff & Yes \\
    CS-12 & Monica Phillips & 10,000 & Common (CS) & \$0.00001 & 1/48th monthly, 1 year cliff & Yes \\
    CS-13 & Christopher Knight & 15,000 & Common (CS) & \$0.00001 & 1/48th monthly, 1 year cliff & Yes \\
    CS-14 & James Smith & 10,000 & Common (CS) & \$0.00001 & 1/12 monthly, no cliff & Yes \\
    CS-15 & Hassan Murphy & 125,000 & Common (CS) & \$0.00001 & 1/48th monthly, 1 year cliff & Yes \\
    CS-16 & Keisha Young & 15,000 & Common (CS) & \$0.00001 & 1/48th monthly, 1 year cliff & Yes \\
    CS-17 & Daniel Brown & 20,000 & Common (CS) & \$0.00001 & 1/48th monthly, 1 year cliff & Yes \\
    CS-18 & Michael Gray & 8,000 & Common (CS) & \$0.00001 & 1/48th monthly, 1 year cliff & Yes \\
    
    \midrule
    \textbf{Grand Total} & & \textbf{8,355,000} & & & & \\
\end{longtable}
\end{landscape}